\documentclass{article}
\usepackage{acra}
\usepackage{amsmath,amssymb}
\usepackage{booktabs}
\usepackage{graphicx}
\usepackage{url}

\title{Sim-to-Real Room-Mediated Object-Goal Navigation with ROS 2 and Nav2}

\author{Matthew Chua \and Adam Scicluna \and Gavin Paul\\
Robotics Institute, Faculty of Engineering and Information Technology\\
University of Technology Sydney, Sydney, Australia}

\begin{document}

\maketitle

\begin{abstract}
Although object-goal navigation (ObjectNav) approaches are usually evaluated in simulation, which offers scale and repeatable environments, real-world deployments are rarely examined beyond feasibility demonstrations. This paper presents the deployment of Room-Mediated Co-occurrence, an ObjectNav policy that reasons over Room Probability Vectors (RPVs), on two robotic platforms (Boston Dynamics Spot and Clearpath Robotics TurtleBot~4) through a robot-agnostic ROS~2 bridge. We study how these semantic priors generalise to physical environments and describe the adaptations required to transfer RPV from simulation to hardware, covering sensing, localisation, and the action interface. Across 66 trials on Spot, searching for 11 object categories from two starting positions in a university postgraduate research facility, RPV achieves a raw Success Rate (SR) of 0.455. Our analysis identifies undetected glass walls, entry into off-limits areas, limited usable depth range, target size, and mismatch between the room lexicon and physical layout as plausible contributors to the 8.4-percentage-point gap between the real-world SR and the reported simulated SR of 0.539. Code is available at \url{https://github.com/Matthew-K-Chua/rpv-ros2-bridge}.
\end{abstract}

\begin{figure}[t]
    \centering
    \includegraphics[width=\linewidth, height=0.35\textheight, keepaspectratio]{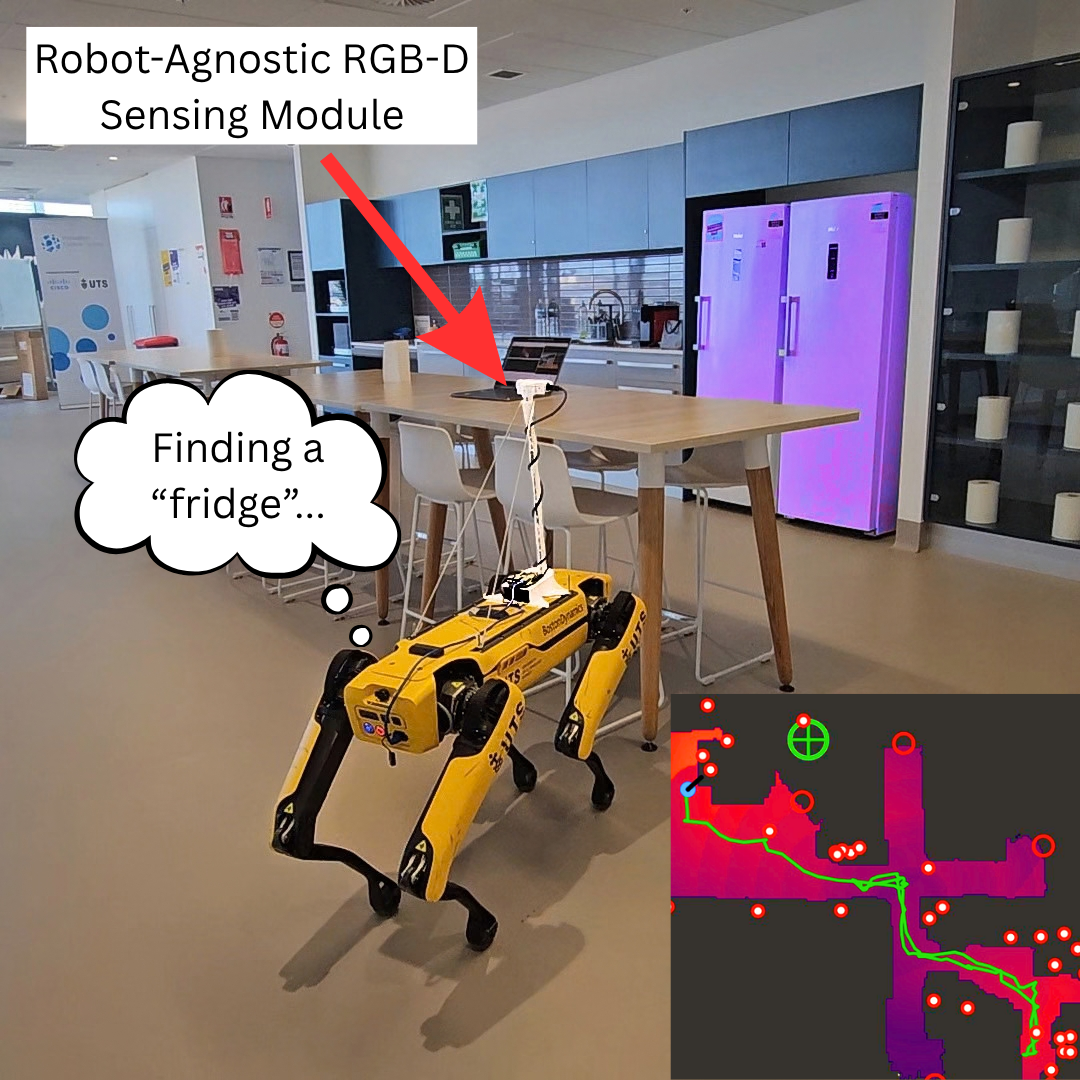}
    \caption{The Boston Dynamics Spot platform searches for the target object ``fridge'' using Room-Mediated Co-occurrence. We mount a custom low-cost on-board ROS~2 sensing module onto Spot that gathers RGB-D data and relays velocity commands to the robot. The green trail outlines the policy's chosen path in this successful trial.}
    \label{fig:fridgeTrial}
\end{figure}
\section{Introduction}
Zero-shot Object-Goal Navigation (ObjectNav)~\cite{ye2021auxiliary} is a foundational problem for general-purpose robotics, requiring an agent to find an instance of a specified object in an initially unseen environment without domain-specific fine-tuning. Large datasets provide the scale and visual fidelity needed to test ObjectNav approaches while abstracting several failure points that become apparent on a real robot, including synchronisation, locomotion, and sensor noise. For example, the RPV simulation benchmark uses 2,000 episodes across 20 HM3D environments~\cite{ramakrishnan2021hm3d} within Habitat~\cite{savva2019habitat,scicluna2026roommediated}. Physical deployment is then needed to establish how the method behaves outside those simulated assumptions.

Recent work on Room-Mediated Co-occurrence~\cite{scicluna2026roommediated} extends Vision-Language Frontier Maps (VLFM)~\cite{yokoyama2024vlfm} with Room Probability Vectors and reports a 3\% relative improvement over the VLFM baseline on the HM3D validation split in simulation. Although VLFM has also been demonstrated on the Boston Dynamics Spot platform, the literature still lacks detailed analysis of how such policies behave and fail when transferred to physical environments.

In this paper, we deploy and evaluate RPV on a Boston Dynamics Spot and demonstrate portability to a Clearpath Robotics TurtleBot~4. Rather than coupling the semantic policy directly to a robot-specific SDK, we communicate through the open-source Robot Operating System~2 (ROS~2)~\cite{macenski2022ros2} and standard navigation interfaces. This design lowers the effort required to evaluate embodied-AI ObjectNav pipelines on different physical robots. Figure~\ref{fig:fridgeTrial} illustrates one successful Spot trial.

The main contributions of this work are as follows:
  \begin{itemize}
      \item A modular ROS~2 bridge implementing RPV on Boston Dynamics Spot and Clearpath Robotics TurtleBot~4.
      \item An analysis of the adaptations required to transfer the simulation pipeline to physical control.
      \item Real-world evaluation and analysis of RPV over 66 trials on Boston Dynamics Spot.
  \end{itemize}

\section{Related Work}
\subsection{Sim-to-Real Reality Gap and Deployment}
Simplified sensing, actuation, and physical interaction in simulation create a divergence from reality such that policies evaluated in simulation may generalise poorly to real-world deployment. Traditional approaches to bridging this ``reality gap'' make an attempt to mimic real life by introducing random sensor noise~\cite{tobin2017domainrandomizationtransferringdeep}, designing high-fidelity physics simulations~\cite{gao2026nvidiaisaacsimenabling}, or using photorealistic datasets such as HM3D~\cite{ramakrishnan2021hm3d}. However, policies benchmarked in simulation can overfit to simulator-specific artefacts, such as taking shortcuts through otherwise non-navigable space~\cite{Kadian_2020}. Truong et al.~\cite{truong2022rethinkingsim2reallowerfidelity} counter-intuitively suggest that lower-fidelity simulation can improve generalisation to real-world navigation.

Despite advances in closing the reality gap, it remains important to evaluate new robotics methods on physical platforms. ROS~2-compatible simulators such as NVIDIA Isaac Sim~\cite{gao2026nvidiaisaacsimenabling} allow engineers to deploy simulated programs onto robots without significant code duplication. Habitat~\cite{savva2019habitat}, one of the most widely used simulators in ObjectNav research, can interface with physical robots through the Habitat-PyRobot Bridge~\cite{Kadian_2020} and the open-source PyRobot framework.

\subsection{Zero-Shot ObjectNav Evaluation on Physical Robotic Platforms}
Recent ObjectNav methods demonstrate strong simulation performance, but their real-world robustness remains largely unverified. Several works evaluate their methods only in simulation~\cite{scicluna2026roommediated,gadre2023cow,zhou2023esc}. Others report successful deployment but do not analyse failure modes or the sim-to-real gap~\cite{yokoyama2024vlfm,ginting2024seek,busch2025onemap}. VLFM~\cite{yokoyama2024vlfm} is deployed through the proprietary Boston Dynamics (BD) SDK, which limits reproducibility and portability to other platforms; SEEK~\cite{ginting2024seek} and OneMap~\cite{busch2025onemap} do not report their deployment frameworks. OpenFrontier~\cite{padilla2026openfrontier} deploys ROS on Spot, reporting an SR of 70\% over 10 trials, but confines its detailed failure analysis to simulation. TopoNav~\cite{liu2025toponavtopologicalgraphskey} deploys on a Unitree Go2 using an Intel RealSense D435i and HESAI XT-6 LiDAR.

Of the deployments that use ROS, L3MVN~\cite{yu2023l3mvn} offers the most detailed account, highlighting the noise that real RGB-D, LiDAR, and odometry data feed into its pipeline. However, it does not examine individual trials or identify which properties of the physical system cause drops in performance. LOAT~\cite{lin2025loat} also reports a successful ROS deployment, but does not provide a per-trial analysis or describe the required sim-to-real adaptations.

\begin{figure*}
    \centering
    \includegraphics[width=0.98\linewidth,trim={1cm 1.7cm 1cm 1.3cm}, clip]{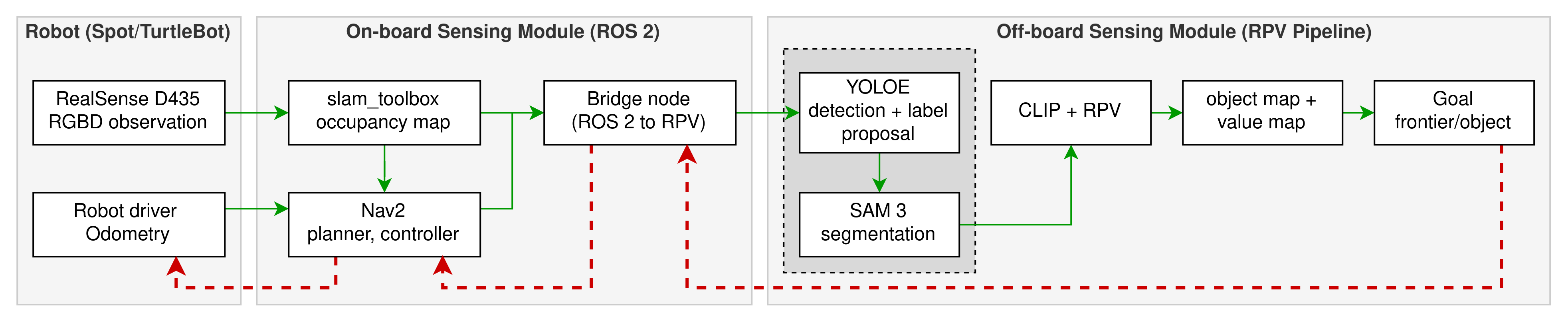}
    \caption{Overview of the system architecture designed to port RPV to a ROS~2 system with an accessible Ethernet port. Green arrows indicate the forward data path, and red dashed arrows indicate the return path.}
    \label{fig:systemArchitecture}
\end{figure*}

\begin{table*}[t]
\centering
\small
\begin{tabular}{lll}
\toprule
Component & Simulation (Habitat) & Physical (Spot) \\
Room lexicon $\mathcal{R}$ & 9 household rooms & 9 experiment-specific labels \\
Stage 1: Mask2Former masking & On & Off \\
Stage 2: YOLOE confidence & $>0.5$ & $>0.3$ \\
Stage 3: SAM~3 confidence & $>0.5 $ & $>0.35$ \\
Perception cadence (explore) & Every 3rd step & Every step \\
Propagation limits & None & Truncated at $3\sigma_o$ and 6\,m geodesic \\
Frontier score $V_j$ & Median of $V(\mathbf{x})$ within 0.5\,m & Maximum within 0.5\,m; empty disc scored 0 \\
Depth range & 5 m (Habitat sensor) & 0.3 to 3.0 m (D435) \\
Obstacle map & Projected from policy depth & Nav2/SLAM occupancy grid at 0.05 m \\
Targets & 6 objects (large, medium) & 11 objects (large, medium, small) \\
\bottomrule
\end{tabular}
\caption{Deviations of our real-world deployment from the benchmarked RPV pipeline.}
\label{tab:sim2real}
\end{table*}

\section{Method}
\subsection{Task Definition}
A ROS~2 robot, initialised from one of two starting poses $P_1$ or $P_2$, must navigate within $d_{\text{success}} = 1.4\,\text{m}$ of the selected target object's location. At each tick $t$, the system receives an RGB snapshot $I_{\text{rgb}}$, a depth raster $I_{\text{depth}}$, an occupancy grid, and an egocentric pose $\xi_t$. Objects segmented by the policy populate the occupancy grid with a propagating co-occurrence signal, creating a semantic value map from which the highest-valued frontier is selected as a navigation goal. The high-level RPV semantic policy then selects a navigation goal $g_t$ for the robot:
\begin{equation}
    g_t = 
    \begin{cases} 
        g_{\text{frontier}}, & \text{if target is undetected} \\
        g_{\text{target}}, & \text{if target is detected}
    \end{cases}
\end{equation}

Nav2~\cite{macenski2020marathon2} then drives the robot towards $g_t$, using NavFn as its global planner to compute a collision-free path on the occupancy map and a Regulated Pure Pursuit controller~\cite{macenski2023regulatedpurepursuitrobot} to track that path by publishing velocity commands to the robot.

An episode ends when one of the following four conditions occurs:
\begin{enumerate}
    \item \textbf{Target Arrival:} The robot reaches the detected target location and transitions to state $S_{stopped}$.
    \item \textbf{Timeout:} The episode reaches the maximum step budget $T_{\max}$ without target detection ($T_{\max} = 250$) (Note that if a target is detected within $T_{\max}$, the episode can extend to beyond 250 ticks to accommodate navigation time).
    \item \textbf{Frontier Exhaustion:} No valid unexplored frontier waypoint remains.
    \item \textbf{Manual Intervention:} An operator issues an emergency stop or halts execution due to a safety breach.
\end{enumerate}

An episode is evaluated as a success if Condition 1 is met and the target object is correctly segmented. Stopping at an incorrect object class is categorised as a false positive. Conditions 2-4 are recorded as failures and categorised by cause.

\subsection{Room Probability Vectors}
Room-Mediated Co-occurrence~\cite{scicluna2026roommediated} is an ObjectNav framework that extends VLFM~\cite{yokoyama2024vlfm} by assigning each detected object spatial context through room labels. An open-vocabulary perception stack segments and localises objects from RGB-D observations and maps each detection to a Room Probability Vector (RPV). The RPV is compared with that of the target, and the resulting semantic signal is written into a value map through geodesic flood fill. Frontiers are ranked on this map, and the highest-valued frontier becomes the next navigation goal.

Several design choices make RPV a practical candidate for physical deployment. The pipeline is modular, so the perception and frontier-scoring process can be detached from agent control. To enable real-time perception, RPV uses a dual perception funnel featuring YOLOE~\cite{wang2025yoloe} for lightweight label proposals, which it then passes to SAM~3~\cite{carion2025sam3} to remove low-likelihood segmentations. The RPV authors advocate robotic deployment and recommend offloading the perception stack to an edge server. We adopt that recommendation directly.

\subsection{Robot-Agnostic Democratisation and Modularity}
Physical demonstrations of zero-shot ObjectNav policies using Boston Dynamics Spot have been programmed directly against its proprietary SDK~\cite{yokoyama2024vlfm,busch2025onemap}. We instead rely on ROS~2 and standard navigation interfaces (laser scans, Nav2, SLAM Toolbox, and velocity commands), confining robot-specific software to the vendor's ROS~2 driver, a Nav2 parameter file, and 126 lines of integration and configuration code for Spot and 45 lines for the TurtleBot~4. The goal is to enable physical evaluation on a wider range of affordable and attainable hardware. For the same reason, we split the computation between an off-board GPU host and a low-cost on-board sensing module that can be moved between platforms, enabling repeatable testing with an identical sensing setup. In exchange for this portability, the system depends on an additional network layer between the GPU host and sensing module.

\subsection{Software Architecture}
Figure~\ref{fig:systemArchitecture} shows a high-level overview of our pipeline. At runtime, the on-board sensing module runs the robot's ROS~2 driver and camera pipeline, projecting a raw 2.5-D depth stream into a 2-D laser scan for mapping and forwarding the depth-aligned RGB-D frames required by the RPV policy. These topics reach the GPU host through a VPN tunnel. On the host, SLAM and Nav2 build an occupancy map and plan motion, and a bridge node sends a snapshot of the robot's occupancy map, pose, and RGB-D observation to the RPV policy over HTTP in the same manner as VLFM. The policy's selected frontier is sent back to Nav2 as the next navigation goal, and the resulting velocity commands are relayed to the robot through the on-board sensing module.

\subsection{Differences Between Simulation and Physical Embodiment}
Deploying RPV on physical robots required adapting the simulation-oriented pipeline in ways that simulation avoids. Table~\ref {tab:sim2real} summarises differences in embodiment, sensing, perception, mapping, navigation, and evaluation. Two key design changes are removing Mask2Former (M2F) and lowering confidence thresholds for the YOLOE and SAM~3 components of the perception stack.

Originally, Mask2Former was used to mask uninformative regions in the RGB image (e.g., floors, ceilings, walls, and other architectural structures). Its computational cost prevented real-time operation on the deployed system, so we removed this stage and instead excluded uninformative labels from the perception vocabulary.

We lowered the YOLOE and SAM~3 confidence thresholds after initial tests produced many false negatives. Motion blur from robot movement and distortion from the D435's rolling-shutter RGB sensor caused prominent objects such as wheelie bins to be missed, leaving too few detections to guide exploration before timeout.

\section{Experimental Setup}
\subsection{Hardware}
\begin{figure}
    \centering
    \includegraphics[width=\linewidth,trim={0 1.5cm 0 0}, clip]{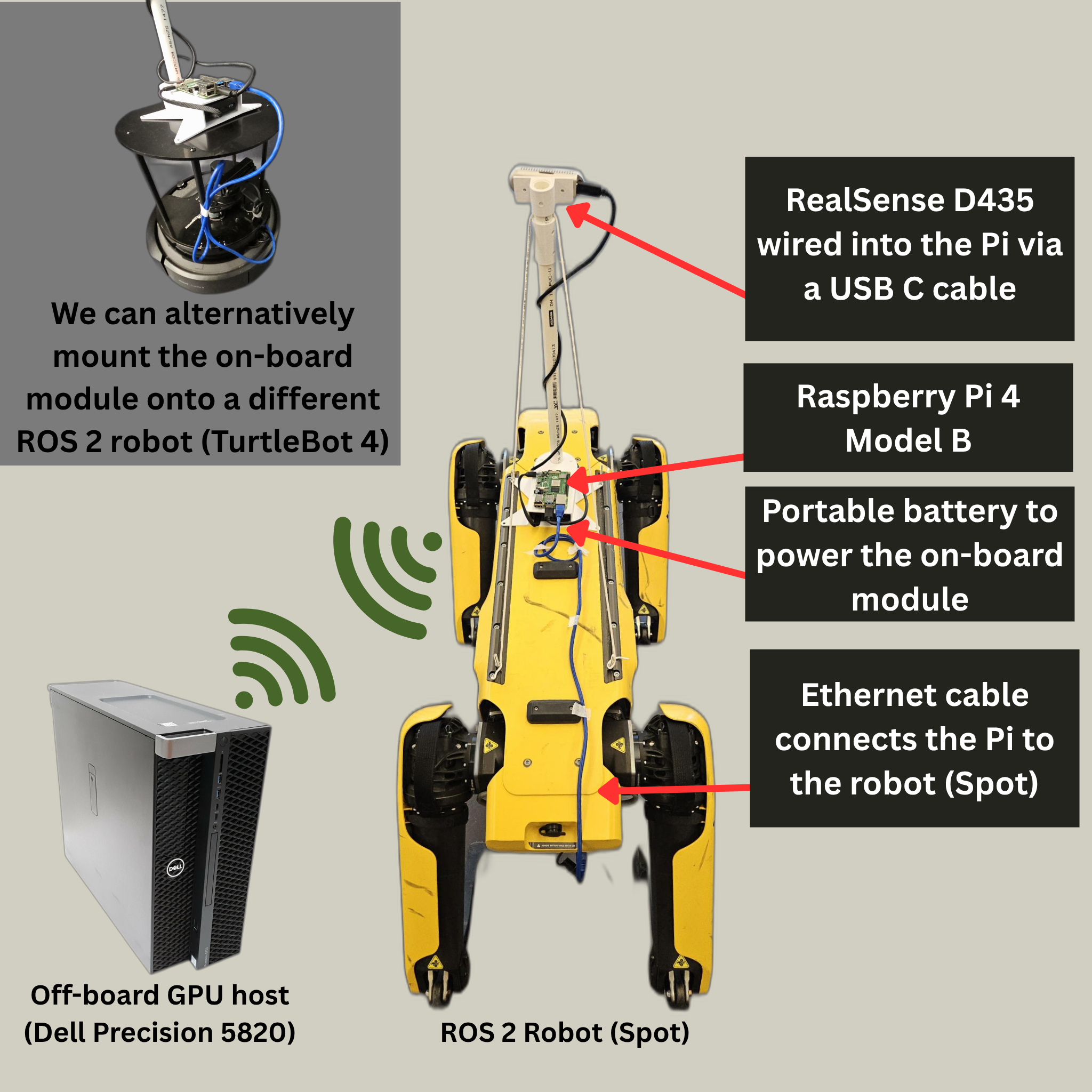}
    \caption{Annotated hardware for Spot and TurtleBot~4. The on-board sensing module comprises a Raspberry Pi~4, RealSense D435, portable battery, and an Ethernet connection to the robot. It communicates with the off-board GPU host over Wi-Fi.}
    \label{fig:AnnotatedHardware}
\end{figure}

Our off-board GPU host is a Dell Precision 5820 equipped with an RTX 3090 and 24~GB of VRAM. It communicates with a low-cost on-board sensing module comprising a RealSense D435 RGB-D camera connected to a battery-powered Raspberry Pi~4 Model~B (Figure~\ref{fig:AnnotatedHardware}). The module is mounted on Spot and secured with a tensioned cord to reduce camera vibration. Experiments are conducted in part of a university floor containing offices, research laboratories, kitchens, an elevator room, and a printing station. We test RPV in an unmodified, uncontrolled environment to evaluate its generalisation to a realistic deployment scenario.

\subsection{Episode Initialisation}
We choose two points in the scene as the robot's starting locations for experimentation. These were deliberately chosen such that during the initialisation phase of each episode, multiple frontiers leading to regions of interest are generated. $P_1$ has three prominent frontiers in the form of hallways that lead to: (a) a printing station, a kitchenette and an elevator room, (b) a robotic research laboratory, and (c) a hallway that leads directly to $P_2$. $P_2$ offers 4 prominent frontiers: (a) the rear entrance to the aforementioned robotic research laboratory, (b) offices, (c) an office meeting space, and (d) a kitchen and an elevator room.

\begin{figure}
    \centering
    \includegraphics[width=0.85\linewidth]{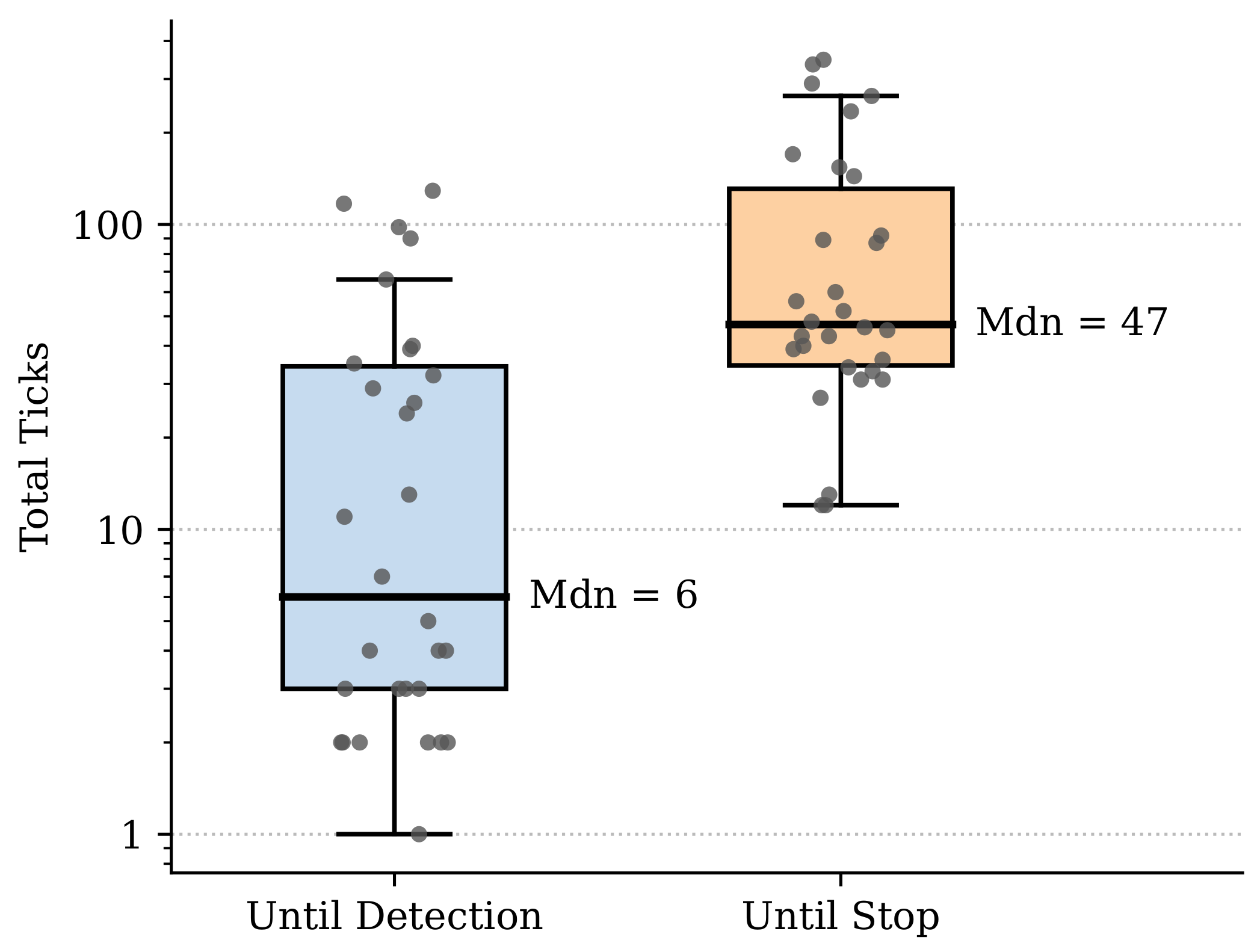}
    \caption{Distribution of policy steps to first detection and stopping over successful physical RPV trials. The vertical axis is logarithmic.}
    \label{fig:avgTimeSteps}
\end{figure}

\subsection{Metrics}
\label{sec:metrics}
To evaluate our physical deployment against simulation benchmarks, we report Success Rate (SR) as a binary metric indicating whether the target was correctly detected and the robot navigated to it. We also measure the policy steps to first target detection, $t_o$, and to the stopping action, $t_s$. To capture failures arising from sim-to-real deployment, we define four categories: timeout (the robot did not detect the target within 250 steps), false positive (the target label was assigned to a non-target object, as verified against ground truth), emergency stop (the operator terminated a run for safety), and banned area (the robot entered a predesignated off-limits area containing none of the selected targets).

\subsection{RPV Variables}
We choose a room lexicon specific to the university environment. The list includes nine terms: meeting area, kitchen, office, elevator room, printing station, laboratory, exit, tooling, and lecture hall. We split our target objects into three size categories (small, medium, and large) to analyse how object size affects the overall search pipeline. Large objects are fridge, printer, garbage bin, and cabinet. Medium objects are projector, microwave, backpack, and desktop monitor. Small objects are laptop, whiteboard marker, and paper towels. For each object, we run six trials: three from $P_1$ and three from $P_2$, giving 66 tests. For consistency with prior Spot deployments~\cite{yokoyama2024vlfm,busch2025onemap}, we perform the controlled experiments exclusively with Spot and use the TurtleBot to demonstrate the portability of the bridge.

\section{Results}
\begin{figure}
    \centering
    \includegraphics[width=0.90\linewidth,trim={0.2cm 0.2cm 0.8cm 0}, clip]{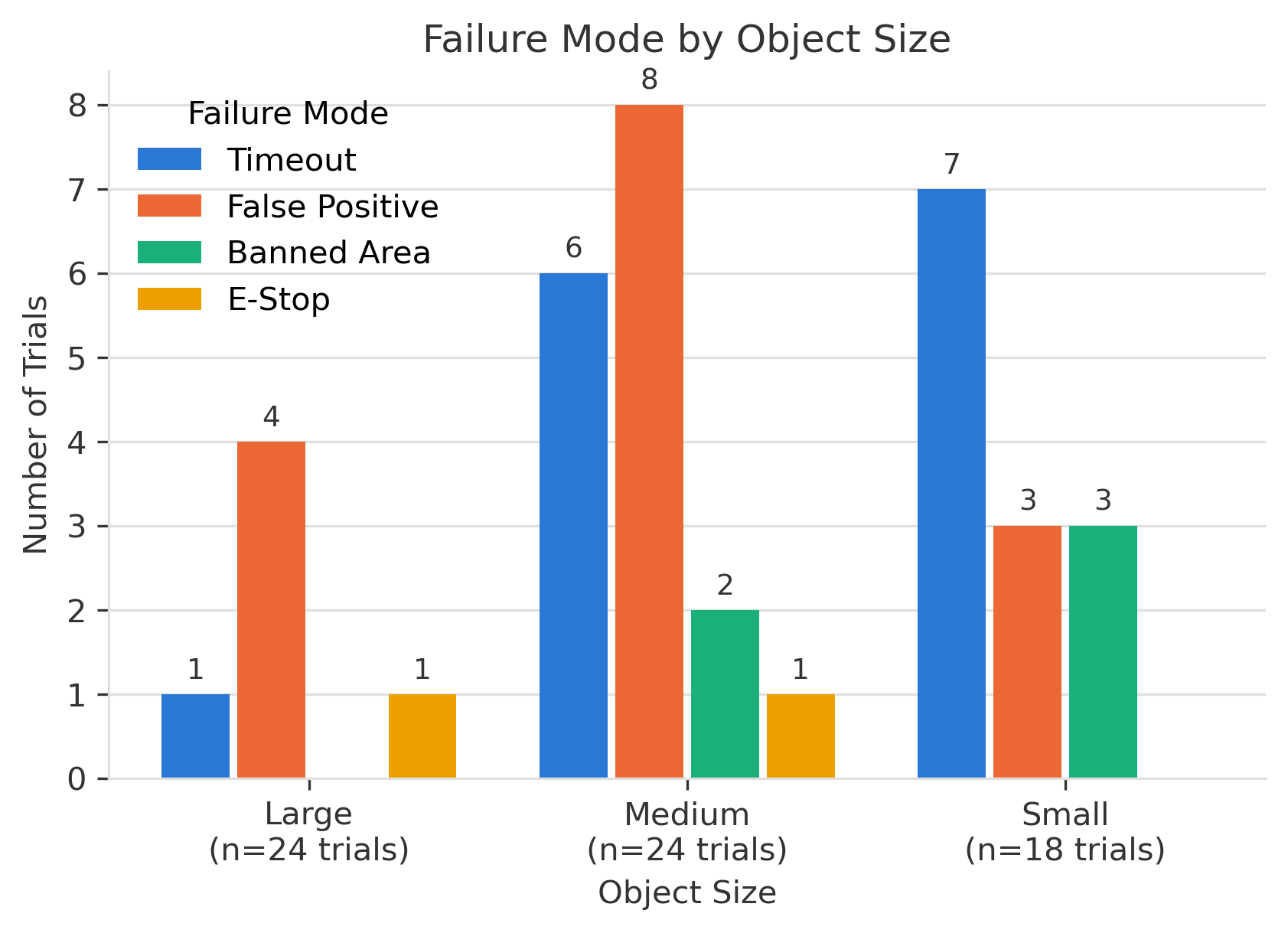}
    \caption{Failure modes aggregated by target-size class.}
    \label{fig:failureBySize}
\end{figure}
Across the 66 trials, the overall SR is 0.455 (30/66). Starting position strongly associates with outcome: $P_1$ achieves an SR of 0.576 (19/33), whereas $P_2$ achieves 0.333 (11/33), a difference of 0.242. Detection time also differs between positions. Successful $P_1$ trials take a mean of 18.3 policy steps to detection (median 3), compared with 40.9 steps (median 35) from $P_2$. Across both starting positions, successful trials take a mean of 26.60 steps to detection (standard deviation 36.72) and 96.93 steps to stop (standard deviation 99.38); Figure~\ref{fig:avgTimeSteps} shows their distributions. Large objects achieve an SR of 0.75, followed by medium and small objects at 0.29 and 0.28, respectively. Table~\ref{tab:results} shows that three targets (microwave, projector, and whiteboard marker) are never found from either position.

Of the four failure modes defined in Sec.~\ref{sec:metrics} and plotted in Figure~\ref{fig:failureBySize}, the 36 failed trials skew toward the medium and small object categories. False positives and timeouts account for 29 failures (15 and 14, respectively). False positives occur in all three size classes but are concentrated among medium objects (8 of 15): the projector produces five and the microwave two. Timeouts are almost exclusively failures on small and medium objects (13 of 14), with the whiteboard marker timing out in five of its six trials.

\begin{table}
\centering
\small
\setlength{\tabcolsep}{1pt}
\begin{tabular}{@{}llcccrr@{}}
\toprule
Size & Target & $P_1$ & $P_2$ & SR & \shortstack{Detect\\($t_o$)} & \shortstack{Stop\\($t_s$)} \\
\midrule
Large  & Cabinet           & 3 & 2 & 0.83 & 14.4 & 37.2 \\
       & Fridge            & 3 & 2 & 0.83 & 40.6 & 93.6 \\
       & Garbage bin       & 3 & 0 & 0.50 &  3.0 & 78.3 \\
       & Printer           & 3 & 2 & 0.83 & 16.6 & 110.4 \\
\addlinespace
Medium & Backpack          & 2 & 0 & 0.33 & 70.0 & 291.0 \\
       & Desktop monitor   & 2 & 3 & 0.83 &  3.2 & 35.4 \\
       & Microwave         & 0 & 0 & 0.00 & --   & --   \\
       & Projector         & 0 & 0 & 0.00 & --   & --   \\
\addlinespace
Small  & Laptop            & 0 & 2 & 0.33 & 58.0 & 230.0 \\
       & Paper towels      & 3 & 0 & 0.50 & 53.0 & 82.7 \\
       & Whiteboard marker & 0 & 0 & 0.00 & --   & --   \\
\midrule
\multicolumn{2}{l}{All trials} & 19/33 & 11/33 & 0.45 & 26.6 & 96.9 \\
\bottomrule
\end{tabular}
\caption{Successes out of three trials per starting position, and the
mean number of policy steps to first detection and to stop over successful trials.}
\label{tab:results}
\end{table}

\begin{figure*}
    \centering
    \includegraphics[width=1\textwidth]{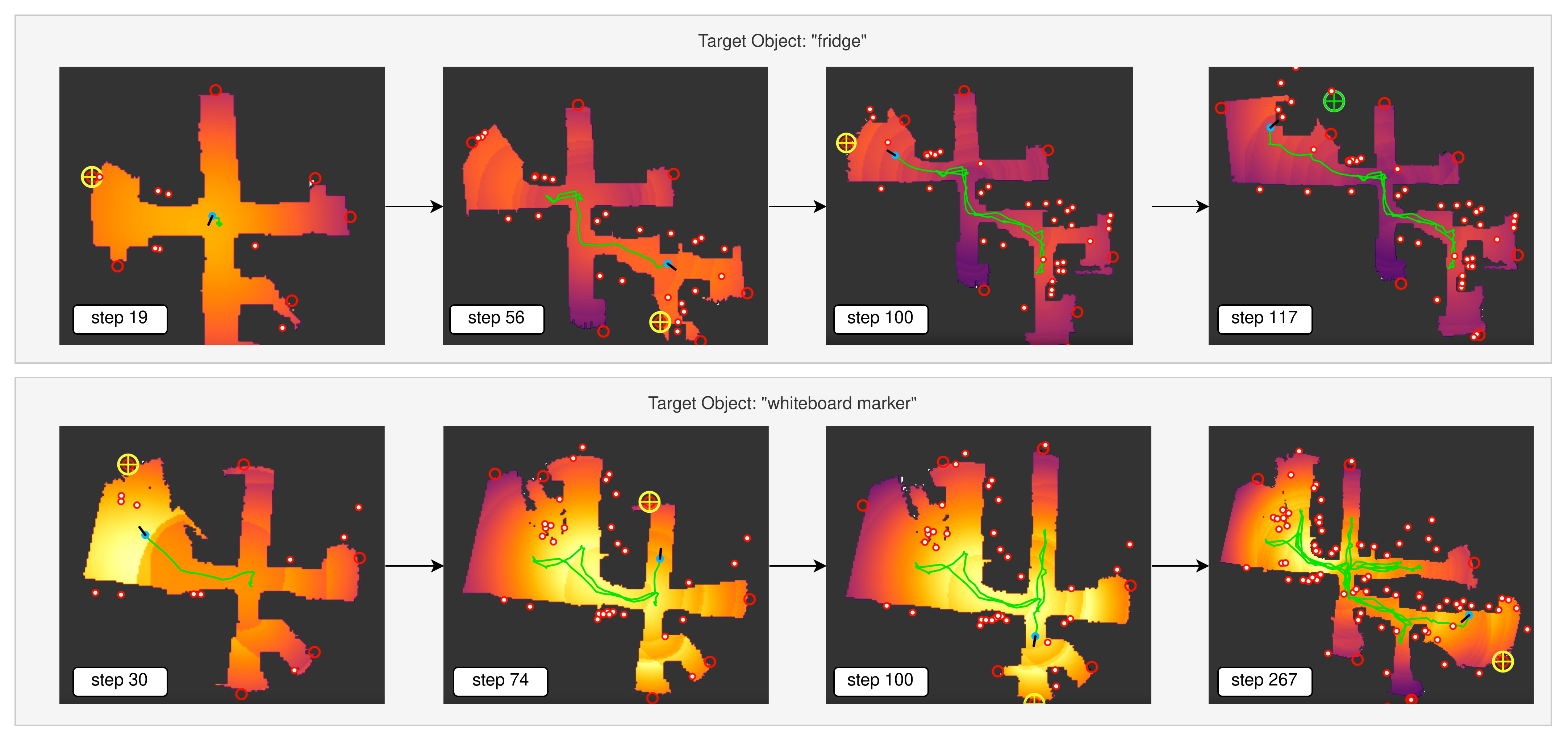} 
    \caption{Chosen frontiers (crosshairs) and paths (green) overlaid on value maps for trial 36, ``fridge'' (top), and trial 62, ``whiteboard marker'' (bottom), at the policy steps labelled in each panel. The fridge search shows kitchen-side object signals eventually outweighing those of the laboratory. The whiteboard-marker search shows a near-flat value map for a high-entropy target and repeated switches between distant frontiers.}
    \label{fig:valueMapPaths}
\end{figure*}

The two remaining failure modes occur only from $P_2$. In five trials, the robot enters a region excluded from the experiment. In two further trials, the operator stops the robot as it attempts to walk through glass. Together, these seven trials account for much of the difference between starting positions. After excluding them, $P_2$'s SR over the remaining 26 trials is 0.423 (11/26), and the pooled adjusted SR is 0.508 (30/59). As a secondary analysis, this narrows the gap to the simulated SR of 0.539 from 8.4 to 3.1 percentage points.

\subsection{Qualitative Behaviour Analysis}
Trial 36 (target object: ``fridge''), shown in the top row of Figure~\ref{fig:valueMapPaths}, demonstrates RPV's intended behaviour. At $t=19$, the robot has begun the ``explore'' phase and sees only generic office objects (locker, television, lamp, and chair), with room co-occurrence values between 0.19 and 0.27. This results in a nearly uniform map, with the best frontier being a hallway leading to the kitchen, worth only 16.1\%. As $t$ grows, the laboratory frontier rises to 22.5\%, and the robot switches headings. After further exploration, at $t=56$, the frontier value shrinks to 15.1\%, and the best frontier switches to the kitchen at 16.5\%. By $t=100$, a chair, laptop, and second locker lift the kitchen-side frontier from 16.9\% to 22.9\% by $t=109$. The fridge is segmented at $t=117$, and Spot stops at $t=264$ after successfully navigating to the target (recall an episode will only end if $T_{\max}$ is reached without target object detection so this trial is still valid). This trial highlights how room co-occurrence scores change as an environment is explored.

By contrast, trial 62 (target object: ``whiteboard marker''), shown in the bottom row of Figure~\ref{fig:valueMapPaths}, shows the consequences of objects with low room association. Every segmented object receives a low room co-occurrence score, with the highest scores at 0.30 for a storage locker and 0.29 for a filing cabinet, while high-entropy furniture contributes little (0.04 for desks and 0.03 for chairs). The best frontier is 9.2\% when the policy enters ``explore'' at $t=12$, and the frontier values decline as $t$ grows. With little contrast, frontier choice is driven by small fluctuations, causing the robot to repeatedly cross the central junction as it switches between distant frontiers at $t=70$, $t=79$, $t=99$. Despite the trial reaching $T_{\max}$ and being recorded as a timeout, we note that at step~264, a pen scores 0.03 and does not redirect the search.

Two limitations follow from this trial. First, the result is consistent with RPV's hypothesis that room co-occurrence is uninformative for high-entropy objects that often move between spaces and are not room-specific. Second, flat value maps can degrade to near-random frontier selection, yielding little information gain per unit distance travelled. In environments with long hallways, such as hospitals or universities, this would be especially costly.

\subsection{Robot-Agnostic Deployment}
We demonstrate robot-agnostic deployment by implementing the same stack used for our Spot experiments on a TurtleBot~4 with limited code changes. We attached our low-cost on-board sensing module to the TurtleBot's top plate. Starting from $P_1$, the robot found an instance of ``paper towels'' at $t=32$ and finished navigating at $t=48$. The TurtleBot's smaller footprint suits the vinyl-floored building, whereas Spot's rectangular body struggled in tight passages between desks in the laboratory. Paths requiring a 180$^{\circ}$ turn with as little as 20~mm of physical clearance caused Nav2 to refuse to plan, contributing to timeouts. Nevertheless, Spot remains the better platform for this feasibility assessment because it handles carpet and small changes in floor elevation that stop the TurtleBot.

Using the original confidence thresholds of 0.50, the policy sometimes failed to segment large objects such as garbage bins in blurred or distorted frames. Lowering the thresholds improved these detections qualitatively. We did not isolate the effects of lighting, sensor noise, or network delay experimentally; after threshold tuning, however, no failure was assigned solely to one of these factors.

\section{Discussion}
Deploying RPV on real hardware exposes strengths and weaknesses that simulation does not. The raw real-world SR is 0.455, or 0.508 after excluding the seven trials terminated by an emergency stop or entry into an off-limits area. We recognise the gap compared with the simulated SR of 0.539~\cite{scicluna2026roommediated} and identify three plausible contributing factors in addition to undetected glass walls and entry into off-limits areas. First, the university layout differs from the simulated scenes: it contains only three semantically clear rooms (kitchen, printing station, and office), while the laboratory, which makes up roughly half of the navigable space, is dominated by high-entropy objects such as chairs and tables, which Scicluna et al.~\cite{scicluna2026roommediated} identify as weak cues. Second, long hallways separate rooms in the test environment, causing costly policy steps to traverse known ground when the best frontier switches to a different hallway and contributing to timeouts for small and medium objects. Finally, the D435's usable depth of about 3\,m results in an occupancy map with a 6\,m diameter during initialisation, compared with a 10\,m diameter in simulation, so frontiers appear closer to the robot and are exhausted quickly, encouraging frontier switching.

Safety interventions play a crucial role in real deployment. Although failed trials arising from entry into off-limits areas reflect experiment-specific conditions rather than unsafe behaviour, glass exposes a significant safety limitation of this sim-to-real deployment. In trials 41 and 52, because the camera did not register an office's hallway-facing glass wall, the policy selected a frontier inside the office and Nav2 planned a path through the glass. During navigation, Spot's internal collision detection also failed to react, and an operator ended each episode before the robot reached the wall. Until transparent surfaces are handled, the system requires close supervision. 

VLFM's real-world Spot deployment integrated the monocular depth estimation model ZoeDepth \cite{bhat2023zoedepthzeroshottransfercombining}, suggesting that learned depth estimation is feasible in physical ObjectNav scenarios. However, metric monocular models still struggle with glass \cite{zheng2026enhancingglasssurfacereconstruction}. Zheng et al. \cite{zheng2026enhancingglasssurfacereconstruction} instead propose a training-free method that aligns an affine-invariant monocular depth prior to raw sensor depth through local RANSAC-based scale-and-shift estimation, recovering glass geometry from the prior. Since the method consumes the existing RGB-D stream, it could run on the GPU host, combining the D435's metric stereo depth with the prior's glass geometry, provided the depth-to-scan projection also moves to the host.

\section{Conclusion}
We deployed Room-Mediated Co-occurrence (RPV) on two ROS~2-compatible platforms, Boston Dynamics Spot and Clearpath Robotics TurtleBot~4, using a robot-agnostic ROS~2-based pipeline for physical ObjectNav evaluation. We analyse how RPV generalises to real-world deployment on an individual-episode basis over 66 Spot trials covering 11 objects and two starting locations. RPV achieved a raw SR of 0.455 compared with 0.539 in simulation, or an adjusted SR of 0.508 after excluding seven safety- or boundary-related terminations. Per-episode analysis shows RPV behaving as intended when the target object has a clear semantic association with a room. The observed gap may be associated with undetected glass walls, off-limits entries, the D435's 3\,m usable range, long hallways with few semantically distinct rooms, and detection thresholds tuned coarsely to reduce false negatives. Two limitations of RPV emerged: high-entropy targets such as a whiteboard marker produced a flat value map, and the frontier hysteresis inherited from VLFM was too weak to prevent repeated retargeting across the building. The D435 also failed to observe glass walls reliably in our environment, creating a safety limitation.

These findings suggest four directions. First, frontier commitment should account for travel cost through distance-aware hysteresis or the information-gain term proposed by the original authors, since repeated switching between hallway ends contributed to timeouts. Second, the effective depth range and 6\,m propagation cap should be evaluated together to determine whether they can be increased without reducing reliability. Third, RPV should be benchmarked in an office or household layout containing rooms that match its lexicon, because our laboratory has little room-level structure. Finally, we do not report Success weighted by Path Length (SPL) because no ground-truth map is available. A geodesic shortest path computed on the final SLAM occupancy grid could provide an estimate for comparison with prior physical deployments~\cite{ginting2024seek}, provided its limitations are reported.

\section*{Acknowledgements}
The authors acknowledge that AI-assisted tools (Claude Opus 5.5, Claude Haiku 4.5, and OpenAI Codex) were used under author direction to plan and draft a manuscript outline from author-provided materials, and for limited text editing. The authors critically reviewed, corrected, and substantially revised all generated material and take full responsibility for the final manuscript. The authors used Gemini Notebook to make the accompanying video. The authors thank UTS:RI for providing the facilities and resources used in this work. The authors also acknowledge support from the Australian Research Council Industrial Transformation Training Centre for Collaborative Robotics in Advanced Manufacturing under grant IC200100001.

{\small
\bibliographystyle{named}
\bibliography{references}
}

\end{document}